\documentclass[11pt]{article}

\usepackage[preprint]{acl}

\usepackage{times}
\usepackage{latexsym}
\usepackage{siunitx}
\usepackage[T1]{fontenc}
\usepackage[utf8]{inputenc}

\usepackage{microtype}
\usepackage{amsmath}
\usepackage{booktabs}
\usepackage{tabularx}
\usepackage{amsfonts}
\usepackage{enumitem}
\usepackage{inconsolata}
\usepackage{listings}

\usepackage{titlesec}
\titlespacing*{\section}{0pt}{5pt plus 1pt minus 1pt}{2pt plus 1pt minus 1pt}
\titlespacing*{\subsection}{0pt}{4pt plus 1pt minus 1pt}{1pt plus 1pt minus 1pt}
\titlespacing*{\paragraph}{0pt}{3pt plus 1pt minus 1pt}{0.5em}

\definecolor{jsonstring}{HTML}{A31515}
\definecolor{jsonkeyword}{HTML}{0451A5}
\lstdefinelanguage{json}{
  morestring=[b]",
  literate=
    *{:}{{{\color{black}{:}}}}{1}
     {,}{{{\color{black}{,}}}}{1}
     {\{}{{{\color{black}{\{}}}}{1}
     {\}}{{{\color{black}{\}}}}}{1}
     {[}{{{\color{black}{[}}}}{1}
     {]}{{{\color{black}{]}}}}{1}
     {true}{{{\color{jsonkeyword}{true}}}}{4}
     {false}{{{\color{jsonkeyword}{false}}}}{5}
     {null}{{{\color{jsonkeyword}{null}}}}{4},
}
\lstdefinestyle{jsonprofile}{
  language=json,
  basicstyle=\ttfamily\footnotesize,
  stringstyle=\color{jsonstring},
  showstringspaces=false,
  breaklines=true,
  breakatwhitespace=false,
  tabsize=2,
  frame=single,
  rulecolor=\color{gray!40},
  numbers=none,
  xleftmargin=2pt,
  xrightmargin=2pt,
  aboveskip=6pt,
  belowskip=6pt,
}

\usepackage{graphicx}

\title{User Preference Modeling for Conversational LLM Agents: Weak Rewards from Retrieval-Augmented Interaction}

\author{{\bf Yuren Hao}, {\bf Shuhaib Mehri}, {\bf ChengXiang Zhai}, {\bf Dilek Hakkani-T\"{u}r} \\
  University of Illinois at Urbana-Champaign \\
  \texttt{\{yurenh2, mehri2, czhai, dilek\}@illinois.edu}}

\usepackage{hyperref}

\begin{document}
\maketitle
\begin{abstract}

Large language models are increasingly used as personal assistants, yet most lack a persistent user model, forcing users to repeatedly restate preferences across sessions. We propose Vector-Adapted Retrieval Scoring (VARS), a pipeline-agnostic, frozen-backbone framework that represents each user with long-term and short-term vectors in a shared preference space and uses these vectors to bias retrieval scoring over structured preference memory. The vectors are updated online from weak scalar rewards from users' feedback, enabling personalization without per-user fine-tuning. We evaluate on \textsc{MultiSessionCollab}, an online multi-session collaboration benchmark with rich user preference profiles, across math and code tasks. Under frozen backbones, the main benefit of user-aware retrieval is improved interaction efficiency rather than large gains in raw task accuracy: our full VARS agent achieves the strongest overall performance, matches a strong Reflection baseline in task success, and reduces timeout rate and user effort. The learned long-term vectors also align with cross-user preference overlap, while short-term vectors capture session-specific adaptation, supporting the interpretability of the dual-vector design. Code, model, and data are available at \url{https://github.com/YurenHao0426/VARS}.
\end{abstract}

\section{Introduction}

LLMs are increasingly used as personal assistants for writing, analysis, and programming \citep{NBERw34255}. During these interactions, users exhibit distinct preferences for how they communicate, receive feedback, and structure workflows \citep{jiang2025personamem,jiang2025personamemv2}, yet most systems lack mechanisms to remember and adapt to these preferences across sessions, forcing users to repeatedly restate them and reducing the efficiency of long-term collaboration \citep{wu2024longmemeval,li2025ldagent, mehri2026multisessioncollab}.

A growing body of work has explored retaining information across sessions for conversational agents. Most systems store past exchanges in a database and retrieve relevant entries at query time, or produce post-session reflections that are reinserted into future conversations \citep{zhong2024memorybank,packer2023memgpt,tan2025prospect,chhikara2025mem0}. Several recent works have also explored using memory to directly improve personalization across sessions \citep{mehri2026multisessioncollab,he2026memoryarena,li2025ldagent}. The retrieval in these systems typically relies on query similarity alone, without a persistent user state to prioritize which preferences are most relevant to the current context \citep{salemi2024lamp,yang2026plugmemtaskagnosticpluginmemory}.

A central gap remains: existing systems recall past interactions but do not maintain a user representation that improves retrieval of relevant preferences through ongoing interaction.

We address this gap with Vector-Adapted Retrieval Scoring (\textbf{VARS}), a framework that learns a compact dual-vector user state from weak scalar rewards from users' feedback and uses it to bias retrieval over structured preference memory. The framework comprises four learned components---a lightweight preference extraction model that converts dialogue into structured memory cards, a preference memory that indexes these cards for retrieval, a dual-vector user state that separates stable cross-session preferences from transient within-session context, and a reward-driven update mechanism that refines the user vectors from weak scalar feedback---layered on top of three frozen backbones (chat LLM, embedding model, reranker). At inference time, the effective user vector biases retrieval scoring so that the most relevant preferences are surfaced for each query. Because adaptation occurs only through compact user-specific vectors, the framework supports continuous personalization without per-user fine-tuning---a form of \emph{user-centric memory} that organizes stored preferences around the user rather than the task.

We evaluate this framework on \textsc{MultiSessionCollab}, an online multi-session benchmark for long-horizon personalization \citep{mehri2026multisessioncollab}. This benchmark pairs each system with an LLM-based user simulator whose preference profiles are enforced across sessions, making it possible to measure whether a persistent user representation improves behavior over time. We instantiate the framework on math and code tasks using open-source LLMs and compare against context-only, reflection-based, and retrieval-based baselines. 
Our results show that the main benefit of VARS is improved interaction efficiency rather than large gains in raw task success. Compared with a strong Reflection baseline, the full VARS system achieves comparable task success while reducing timeout and user effort, indicating that a persistent user representation helps the agent surface relevant preferences with less corrective interaction. More broadly, these findings suggest a distinction between task-centric and user-centric memory objectives: task-centric memory is naturally reflected in whether the task is eventually completed, whereas user-centric memory is also reflected in the cost of collaboration required to reach that outcome. In this setting, organizing memory around the user primarily makes collaboration more efficient, while still yielding modest improvements in task success.

The contributions of our work are:
\begin{itemize}[nosep,leftmargin=*]
\item We propose VARS, a frozen-backbone framework that learns a per-user dual vector from weak interaction feedback and adds a learned user-aware bonus to the task-centric reranker, so that retrieval reflects both query relevance and individual user preference.
\item We evaluate VARS on \textsc{MultiSessionCollab} and show that it improves interaction efficiency---reducing timeout rate and user effort---while matching a strong baseline in task success.
\item We analyze the learned user vectors and show that the dual-vector design separates stable cross-session preferences from transient within-session context, with long-term vectors aligning with cross-user preference overlap.
\end{itemize}

\section{Related Works}

\paragraph{LLM personalization.}
Prior work on LLM personalization includes profile-augmented prompting \citep{salemi2024lamp}, learned user representations \citep{jiang2025personamemv2}, and retrieval-based profile optimization \citep{du2026purple}. These approaches have established the importance of adapting LLM behavior to individual users, but they typically assume a fixed user profile provided as input rather than one learned and updated from ongoing interaction. In contrast, our method learns a compact user representation online from interaction feedback and uses it to guide retrieval over structured preference memory, without per-user fine-tuning.

\paragraph{Long-term memory and multi-session user modeling.}
A closely related line of work equips LLM agents with long-term memory by storing dialogue history, retrieved notes, or reflection summaries for later use \citep{zhong2024memorybank,packer2023memgpt,tan2025prospect,chhikara2025mem0,yang2026plugmemtaskagnosticpluginmemory,sarin2025memoria}. Some systems further separate long-term and short-term state through modular or event-centric architectures \citep{li2025ldagent,zou2026esmem}. Related benchmarks study long-horizon memory and user modeling across sessions through profile inference, memory probing, or multimodal memory tasks \citep{jiang2025personamem,jiang2025personamemv2,wu2024longmemeval,bei2026memgallery,shen2026evolmem}, while \citet{hu2026opbench} highlight the complementary risk of over-personalization. Among these, \textsc{MultiSessionCollab} is closest to our setting because it evaluates downstream collaboration under persistent user preferences rather than question answering over stored history \citep{mehri2026multisessioncollab}. Building on this setting, we ask whether a learned persistent user state improves collaboration over time by reducing user effort and corrective interaction, not only by increasing eventual task success.

\paragraph{Learning from weak interaction feedback.}
Our online update mechanism is also related to learning from implicit or bandit-style feedback in ranking and preference elicitation \citep{rendle2009bpr,hu2008implicitcf,koren2009matrix,zhao2022banditpref}. \citet{du2026purple} optimize profile selection via contextual bandits with a global policy; in contrast, we keep all backbone models frozen and learn a per-user retrieval bias from weak scalar feedback.

\section{Method}

\label{sec:method}

\begin{figure*}[!t]
  \centering
  \includegraphics[width=0.85\textwidth]{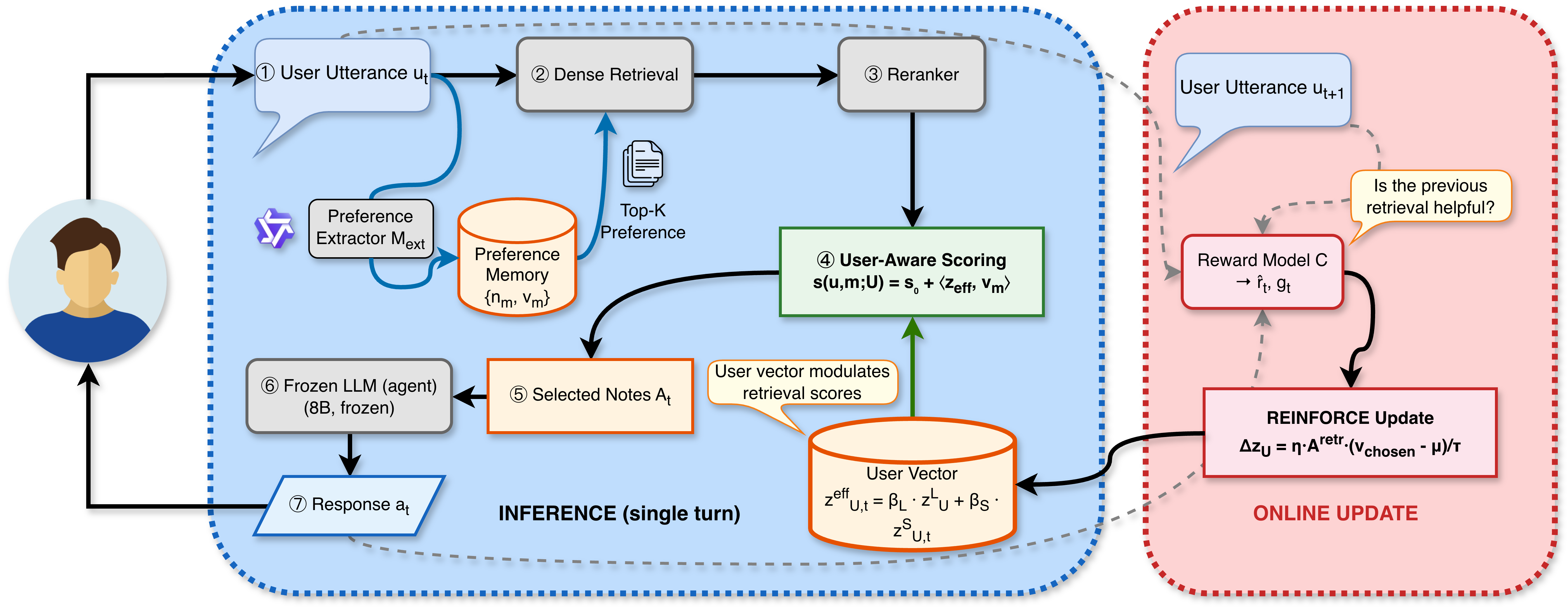}
  \caption{VARS architecture. $M_{\text{ext}}$ extracts preference cards; a dual user vector $z_{U,t}^{\mathrm{eff}} = \beta_L z_U^{(L)} + \beta_S z_{U,t}^{(S)}$ biases reranking; a keyword-based reward $\hat r_t$ drives REINFORCE updates.}
  \label{fig:architecture}
\end{figure*}

Our goal is to learn a persistent representation of the user that improves through interaction and enables the system to selectively surface relevant preferences at inference time. This representation is low-dimensional, modulates retrieval and reranking over structured preference memory, and requires no modification to backbone model parameters, and captures preferences that arise from user interaction as structured condition--action rules.

\subsection{Problem Setting}
\label{sec:method-problem}

A user $U$ interacts with an assistant across sessions $s \in \{1, \dots, S_U\}$. Each session $s$ is a conversation $C_s = \{(u_1, a_1), \dots, (u_{T_s}, a_{T_s})\}$ of $T_s$ turns, where $u_t$ and $a_t$ are the user and assistant utterances at turn $t$. A user utterance may pose a task query (e.g., a math problem) and simultaneously reveal preferences. The assistant must produce responses that are both task-correct and preference-aligned. User preferences persist across sessions, and the system maintains a per-user state $(z^{(L)}_U, z^{(S)}_{U,t})$ between them.

\subsection{System Overview}
\label{sec:method-overview}

Our framework uses three frozen backbone components---a chat model $M_{\text{chat}}$ for response generation, an embedding model $f_{\text{emb}}$ for encoding preference notes, and a reranker $M_{\text{rerank}}$ for scoring query--memory pairs---and introduces four learned components: a lightweight preference extraction model that converts dialogue into structured memory cards, a preference memory that indexes these cards for retrieval, a dual-vector user state that separates stable cross-session preferences from transient within-session context, and a reward-driven update mechanism that refines the user vectors from weak scalar feedback. All backbone parameters remain fixed; adaptation occurs only through the user-specific state.

As illustrated in Figure~\ref{fig:architecture}, the system performs the following steps for a user $U$:
\begin{enumerate}[nosep]
  \item \textbf{Preference Extraction.}
    A small finetuned model $M_{\text{ext}}$ processes a window
    of recent turns and extracts structured
    $(\text{condition}, \text{action})$ preference tuples in a
    fixed JSON schema (e.g., condition: ``solving algebra'',
    action: ``show each step'').
  \item \textbf{Preference Memory.}
    Each extracted tuple is stored as a \emph{memory card}
    comprising the tuple itself, a natural-language note $n_m$,
    a global/conditional flag, and a dense embedding $e_m$ from
    $f_{\text{emb}}$ (schema and example in
    Appendix~\ref{app:card-example}).
  \item \textbf{User Vector.}
    Memory embeddings are mapped into a shared item space;
    a long-term vector $z^{(L)}_U$ is computed as the mean of
    a user's item vectors, and a short-term vector
    $z^{(S)}_{U,t}$ is updated online from the same scalar
    reward signal (Section~\ref{sec:method-rl}).
  \item \textbf{Personalized Retrieval.}
    For query $u_t$, candidate memory cards are retrieved
    by dense search, reranked with a cross-encoder, and
    adjusted by a user-dependent bonus from
    $z^{\text{eff}}_{U,t} = \beta_L z^{(L)}_U + \beta_S z^{(S)}_{U,t}$. Universally applicable
    preferences bypass retrieval and are injected directly
    (Section~\ref{sec:method-memory}).
  \item \textbf{Online RL Update.}
    The follow-up query $u_{t+1}$ is converted into a scalar
    reward via keyword matching (Section~\ref{sec:method-rl});
    a REINFORCE-style update then adjusts $z^{(L)}_U$ and
    $z^{(S)}_{U,t}$.
\end{enumerate}

\subsection{Preference Extraction Model}
\label{sec:method-extractor}

We use $M_{\text{ext}}$, a 0.6B-parameter Qwen3 model~\cite{yang2025qwen3}, as a lightweight preference extractor. The model is finetuned for instruction following and structured JSON generation. To train it, we construct a dataset of $564$K examples by combining public chat logs (LMSYS-Chat, WildChat), instruction-tuning corpora (Alpaca, SlimOrca), and GPT-5.1--labeled preference JSON~\cite{zheng2023lmsyschat1m,zhao2024wildchat,wang2023camels}. On a held-out set, $M_{\text{ext}}$ achieves $99.7\%$ JSON validity and $97.5\%$ recall at $37.7\%$ precision: it over-extracts, but the downstream reranker and user vector filter irrelevant cards, making high recall the priority. A per-source breakdown is in Appendix~\ref{app:extractor}, Table~\ref{tab:extractor-data}. The trained model and training data are publicly available.\footnote{\url{https://huggingface.co/blackhao0426/pref-extractor-qwen3-0.6b-full-sft}; \url{https://huggingface.co/datasets/blackhao0426/user-preference-564k}}

\subsection{Preference Memory and Item Space}
\label{sec:method-memory}

\paragraph{Memory cards.}
For each dialogue window $W_t$ with non-empty extracted preferences $P_t$, we instantiate one or more memory cards. A memory card $m$ for user $U$ stores the user and session identifiers, the source turn IDs and raw user queries $\{u_k^{(m)}\}$, a preference subset $P_m \subseteq P_t$, a short textual note $n_m$ summarizing the extracted preference(s) (e.g., ``When doing code generation, use Python.''), and an embedding $e_m = f_{\text{emb}}(u_{\text{source}}^{(m)}) \in \mathbb{R}^d$ ($d = 4096$) of the source user query that expressed the preference.

\paragraph{Global and conditional preferences.}
Not all extracted preferences need to be retrieved at inference time. Some are globally applicable (e.g., ``always respond in Chinese''), whereas others are conditional on task type or local context (e.g., ``when coding, include type hints''). We therefore classify each extracted preference as \emph{global} or \emph{conditional} from its condition field. A preference is treated as global if its condition contains universal indicators such as ``general,'' ``always,'' or ``any task,'' or if it consists of fewer than three words and contains no domain-specific terms. Global preferences bypass retrieval and are injected directly into the agent prompt, up to a cap of $10$, while only conditional preferences enter the dense retrieval and reranking pipeline. This two-tier design preserves retrieval capacity for query-dependent preferences while ensuring frequently applicable preferences are not missed; misclassified global preferences can still be surfaced through the conditional retrieval pipeline, limiting the impact of classification errors.

\paragraph{Shared item space via PCA.}
We project memory embeddings into a shared lower-dimensional item space via PCA. Given all memory embeddings $\{e_m\}_{m=1}^{M}$ with mean $\mu$, we retain the top $k$ principal components in $P \in \mathbb{R}^{k \times d}$ and define the item vector $v_m = P(e_m - \mu) \in \mathbb{R}^k$. We set $k = 256$ to reduce the cost of user-vector operations while retaining the dominant structure of the embedding space~\citep{kusupati2022matryoshka,khaledian2025pcarag}. This shared coordinate system enables direct comparison between memory cards and learned user vectors (Section~\ref{sec:method-user-vector}).

\subsection{User State and User-Centric Retrieval}
\label{sec:method-user-vector}

\paragraph{Long-term and short-term user vectors.}
For each user $U$, we maintain two learned vectors in
$\mathbb{R}^k$. The \emph{long-term vector}
$z^{(L)}_U \in \mathbb{R}^k$ is initialized to zero at the start
of the first session and updated across all subsequent sessions
via the REINFORCE-style rule in
Section~\ref{sec:method-rl}. Because it is never reset, it
accumulates information from past interactions and is intended to
capture stable cross-session preferences, such as preferred
language or level of detail. The \emph{short-term vector}
$z^{(S)}_{U,t} \in \mathbb{R}^k$ is initialized to zero at
the start of each session, updated from turn-level feedback, and
exponentially decayed to down-weight older signals. It is
intended to capture transient within-session context and recency
effects. The effective user vector at turn $t$ is
\[
  z^{\mathrm{eff}}_{U,t}
  = \beta_L z^{(L)}_U + \beta_S z^{(S)}_{U,t},
\]
where $\beta_L, \beta_S \ge 0$ control the relative influence of
cross-session and within-session state. Both vectors are learned
from interaction feedback alone; no explicit user features or
pre-computed user centroids are required.

\paragraph{Base reranker.}
Given query $u_t$ and candidate memories
$M_t = \{m_1,\dots,m_K\}$ from dense retrieval, the frozen
reranker $M_{\text{rerank}}$ computes base relevance scores
\[
  s_0(u_t,m_i)
  = \log p_{\text{rerank}}(y=1 \mid u_t, n_{m_i}),
\]
where $n_{m_i}$ is the textual note of memory $m_i$. These scores
depend only on the query and note text and are independent of the
user state.

\paragraph{User-aware scoring.}
We add a user-specific bonus based on the dot product between the
effective user vector and the memory-card item vector:
\[
  s(u_t,m_i;u)
  = s_0(u_t,m_i) + \langle z^{\mathrm{eff}}_{U,t}, v_{m_i} \rangle.
\]
This yields a low-rank residual scoring layer on top of the
frozen reranker, linear in both $z^{\mathrm{eff}}_{U,t}$ and
$v_{m_i}$, and introduces no new global parameters beyond the
user vectors. We define a softmax retrieval policy over the
candidate set:
\[
  \pi_z(m_i \mid u_t,U)
  = \frac{\exp(s(u_t,m_i;u)/\tau)}
         {\sum_{j=1}^{K} \exp(s(u_t,m_j;u)/\tau)},
\]
with temperature $\tau > 0$. In practice, we inject the top-$J$
memories under $s(\cdot)$ into the LLM prompt.

To bridge the semantic gap between task-oriented queries and preference descriptions during dense retrieval, we apply a lightweight keyword-based query transformation (Appendix~\ref{app:query-transform}).

\subsection{Weak Reward Instantiation and Online User-State Update}
\label{sec:method-rl}

A central design choice of the framework is that user-state learning depends only on a scalar feedback signal. The update rule itself is agnostic to how this signal is produced; in our experiments, we instantiate it with a lightweight heuristic computed from the user's next turn $u_{t+1}$. Intuitively, this follow-up turn provides weak evidence about local collaboration quality: if the user continues without restating preferences or correcting the response, the preceding interaction is treated as more favorable; if the user expresses dissatisfaction or repeats preference constraints, it is treated as less favorable.

\paragraph{Keyword-based reward estimation.}
We compute a scalar reward $\hat r_t$ from $u_{t+1}$ using two lightweight signals:
\begin{enumerate}[nosep]
  \item \emph{Sentiment keywords.}
    Curated negative indicators (e.g., ``incorrect'', ``redo'') contribute $-1.0$, while positive indicators (e.g., ``thanks'', ``continue'') contribute up to $+1.0$.
  \item \emph{Topic coherence.}
    Cosine similarity between $e_{u_t}$ and $e_{u_{t+1}}$ is used to detect topic shifts. When the similarity is low ($< 0.2$), the reward is dampened, since it becomes less clear whether the follow-up turn reflects satisfaction with the previous response.
\end{enumerate}
The final reward is clipped to $[-1,1]$. This instantiation requires no additional model call; a sensitivity analysis is in Appendix~\ref{app:sensitivity}.

\paragraph{Heuristic retrieval-attribution gate.}
Because $\hat r_t$ may reflect generation failures unrelated to retrieval, we scale the update by a gating factor $g_t \in [0,1]$ that heuristically attributes reward to the retrieval decision. Let
\[
  s_q^{\max} = \max_i \cos(e_{m_i}, e_{u_t})
\]
denote the maximum similarity between the query and any retrieved memory. We then use the following rule:
\begin{itemize}[nosep,leftmargin=*]
  \item Strongly negative reward and no similar retrieved memory ($s_q^{\max} < 0.2$): assign high retrieval attribution, $g_t = 0.9$.
  \item Strongly negative reward and a relevant retrieved memory ($s_q^{\max} > 0.5$): assign low retrieval attribution, $g_t = 0.2$.
  \item Positive reward with a relevant retrieved memory: $g_t = 0.6$; otherwise $g_t = 0.3$.
\end{itemize}
A sensitivity analysis on the ablation logs (Appendix~\ref{app:sensitivity}) confirms that the learned vectors are robust to moderate reward perturbations but that gating is critical: removing it causes severe vector inflation and directional drift.

We maintain a running per-user baseline
\[
  b_U \leftarrow (1-\alpha)b_U + \alpha \hat r_t,
\]
with smoothing coefficient $\alpha$, following standard variance reduction practice for policy gradients~\citep{williams1992simple,greensmith2001variance}.

\paragraph{REINFORCE-style user-state update.}
Let $A_t \subseteq M_t$ denote the memories actually injected at turn $t$. We define the average item vector of the selected memories and the policy expectation:
\begin{equation}
\begin{aligned}
v_{\mathrm{chosen},t}
&= \frac{1}{|A_t|}\sum_{m_i \in A_t} v_{m_i}, \\
\mu_t
&= \sum_{i=1}^{K} \pi_z(m_i \mid u_t,U)\, v_{m_i}.
\end{aligned}
\end{equation}
The retrieval-specific advantage is
\begin{equation}
  A_t^{\mathrm{retr}} = g_t(\hat r_t - b_U),
\end{equation}
which down-weights updates when retrieval is unlikely to be responsible for the observed reward or when the reward is close to baseline.

The long-term and short-term update increments are
\begin{equation}
\begin{aligned}
\Delta z_U^{(L)}
&= \eta_L \frac{A_t^{\mathrm{retr}}}{\tau}\bigl(v_{\mathrm{chosen},t} - \mu_t\bigr), \\
\Delta z_{U,t}^{(S)}
&= \eta_S \frac{A_t^{\mathrm{retr}}}{\tau}\bigl(v_{\mathrm{chosen},t} - \mu_t\bigr),
\end{aligned}
\end{equation}
with learning rates $\eta_L,\eta_S$. We then update the long-term state by
\begin{equation}
  z_U^{(L)} \leftarrow z_U^{(L)} + \Delta z_U^{(L)},
\end{equation}
and the short-term state by
\begin{equation}
  z_{U,t+1}^{(S)} = (1-\lambda) z_{U,t}^{(S)} + \Delta z_{U,t}^{(S)},
\end{equation}
where $\lambda \in (0,1)$ is the decay rate~\citep{besbes2015nonstationary}. Positive advantage moves the user state toward the selected memories, while negative advantage pushes it away.

Because user vectors and memory-card vectors share the same item space, repeated updates toward similar retrieved preferences can lead users with similar revealed preferences to occupy nearby regions of that space, analogous to implicit collaborative filtering~\citep{hu2008implicitcf,koren2009matrix}. We test this empirically in Section~\ref{sec:results}. A theoretical motivation for the update dynamics is provided in Appendix~\ref{app:theory}. In brief, we show that the implemented updates correspond to exact gradients of a fixed-candidate surrogate objective (Proposition~1), and that the dual-vector state admits an exact two-timescale decomposition separating persistent signal from transient local context (Proposition~2).

\subsection{Inference and Adaptation Loop}
\label{sec:method-pipeline}

% NOTE: Figure commented out — duplicate of fig:architecture above.
% \begin{figure*}[t]
%   \centering
%   \includegraphics[width=\textwidth]{latex/sections/pipeline.png}
%   \caption{Overall architecture of the frozen-backbone user modeling framework. Preference extractor $M_{\text{ext}}$ converts dialogue windows $W_t$ into structured preference cards stored in a preference memory with item vectors $v_m$. A learned per-user state consisting of a long-term vector $z_U^{(L)}$ and a short-term vector $z_{U,t}^{(S)}$ is combined into $z_{U,t}^{\mathrm{eff}}$ and biases retrieval and reranking when selecting preference notes $A_t$ to condition the frozen chat LLM $M_{\text{chat}}$. A reward model $C$ maps interaction signals $(u_t, a_t, u_{t+1}, A_t)$ into a scalar reward $\hat r_t$ and gate $g_t$, which drive online updates of the user state.}
%   \label{fig:pipeline}
% \end{figure*}

Figure~\ref{fig:architecture} shows how the components described above interact at each turn. When a new query $u_t$ arrives, the system retrieves candidate preference cards from memory via dense search over $f_{\text{emb}}$ embeddings, then reranks them using $M_{\text{rerank}}$ augmented by the user-dependent bonus from $z_{U,t}^{\mathrm{eff}}$. The top-ranked preference notes, together with any global preferences, are injected into the prompt for $M_{\text{chat}}$, which generates the response $a_t$. Once the user's follow-up $u_{t+1}$ is observed, the reward model produces $\hat{r}_t$ and the attribution gate $g_t$, driving a REINFORCE-style update of both user vectors. In parallel, $M_{\text{ext}}$ processes recent dialogue windows to extract new preference cards that are added to memory. This creates a closed loop: the preference memory grows with interaction, and the user state adapts to steer retrieval toward increasingly relevant preferences. At session boundaries, $z_{U,t}^{(S)}$ resets while $z_U^{(L)}$ persists, separating transient context from stable cross-session signal.

\section{Experimental Setup}

\label{sec:setup}

% \subsection{Benchmark and Tasks}
% \label{sec:setup-benchmark}

% We evaluate our framework on the \textsc{MultiSessionCollab}
% benchmark, which provides a multi-session user-simulation
% environment with LLM-based user simulators and rich
% persona-style preference profiles.
% We select three task domains that require both correctness
% and style compliance:
% %
% \begin{itemize}
%   \item \textbf{math-hard}: complex mathematical problems with
%         ground-truth \LaTeX{} solutions;
%   \item \textbf{math-500}: problems from the MATH-500 benchmark,
%         covering a broad range of mathematical topics;
%   \item \textbf{bigcodebench}: code-generation tasks with
%         executable test cases.
% \end{itemize}
% %
% In all domains, the agent must both produce a correct solution
% and respect the user's style preferences (e.g., level of detail,
% formatting, language) across multiple sessions.

% We run a large-scale evaluation with $60$ user profiles.
% Each profile is run for $60$ sessions, with each session
% allowing up to $10$ interaction turns, yielding up to
% $36{,}000$ sessions in total per method.
% All results in Section~\ref{sec:results} are reported on this
% full set of profiles and sessions.

\subsection{Benchmark and Tasks}
\label{sec:setup-benchmark}

We evaluate on \textsc{MultiSessionCollab}~\citep{mehri2026multisessioncollab}, a benchmark that pairs each system with an LLM-based user simulator whose persona encodes a rich set of style preferences. Evaluation proceeds across multiple sessions per user profile, so the central challenge is whether the system can learn and leverage user preferences over successive interactions rather than treating each session independently. We select three task domains requiring both correctness and style compliance: \textbf{math-hard} (complex problems with ground-truth \LaTeX{} solutions), \textbf{math-500} (broad mathematical topics), and \textbf{bigcodebench} (code generation with executable test cases). In all domains the agent must produce a correct solution while respecting the user's style preferences across sessions.

For each system mode, we evaluate the same $60$ user profiles over $60$ sessions per profile, with up to $10$ turns per session, yielding $3{,}600$ sessions per method. Reported aggregate metrics are computed over all sessions pooled across the three domains.

% \subsection{Profiles and Style Preferences}
% \label{sec:setup-profiles}

% Each user profile encodes a rich set of style preferences as
% condition--action rules.
% Across the $60$ evaluation profiles, the average number of
% preferences per profile is $43$, covering preferences about,
% for example, how to present algebraic derivations, how much
% intermediate reasoning to show, which language to use, and
% how to format code output.

% Preferences are stored in a structured JSON schema with explicit
% identifiers, conflict groups, and priority cues; each preference
% specifies when it applies (condition) and what the agent should do
% (action), together with a conflict group and a small list of
% priority-context keywords. A complete example profile is provided
% in Appendix~\ref{app:profiles}.

% In principle, the same schema can encode non-style
% preferences such as topical interests, tool and API choices,
% or safety constraints; in this work we instantiate it for
% style preferences because the available user-simulation
% benchmark provides rich style-oriented profiles and
% automatic evaluation signals.

\subsection{Profiles and Style Preferences}
\label{sec:setup-profiles}

Each user profile specifies style preferences as structured
condition--action rules that persist across sessions. Across the
$60$ profiles, the average profile contains $43$ preferences,
covering dimensions such as algebraic derivation style, degree of
intermediate reasoning, language choice, and code formatting.
Preferences are represented in a structured JSON schema with
explicit identifiers, conflict groups, and priority cues.
Appendix~\ref{app:profiles} provides a complete example profile.

% \subsection{Models and System Implementation}
% \label{sec:setup-models}

% We use an open-source LLM stack throughout.
% The user simulator is
% \texttt{Llama-3.3-70B-Instruct}~\cite{llama3},
% served via \texttt{vLLM}~\cite{kwon2023vllm} with tensor
% parallelism across two H200 GPUs (GPU~0--1).
% The conversational agent is
% \texttt{Llama-3.1-8B-Instruct}~\cite{llama3},
% served via \texttt{vLLM} on a single H200 GPU (GPU~2).
% For preference memory, we use
% \texttt{Qwen3-Embedding-8B}~\cite{zhang2025qwen3embeddingadvancingtext}
% as the dense embedding model and \newline
% \texttt{Qwen3-Reranker-8B}~\cite{zhang2025qwen3embeddingadvancingtext}
% for scoring query--memory pairs; both share GPU~3.
% The reward model for online user-vector updates
% (Section~\ref{sec:method-rl}) uses a keyword-based heuristic
% that classifies the user's follow-up message into reward
% categories without an additional LLM call.

\subsection{Models and System Implementation}
\label{sec:setup-models}

We use an open-source stack throughout. The benchmark user
simulator is \texttt{Llama-3.3-70B-Instruct}~\cite{llama3}, and
the conversational agent is \texttt{Llama-3.1-8B-Instruct}~\cite{llama3},
both served with \texttt{vLLM}~\cite{kwon2023vllm}. For
preference memory, we use
\texttt{Qwen3-Embedding-8B}~\cite{zhang2025qwen3embeddingadvancingtext}
for dense retrieval and
\texttt{Qwen3-Reranker-8B}~\cite{zhang2025qwen3embeddingadvancingtext}
for query--memory scoring. Preference extraction uses the
lightweight finetuned model $M_{\text{ext}}$ described in
Section~\ref{sec:method-extractor}. All backbone components are
kept frozen during evaluation; online adaptation occurs only
through the user vectors. The scalar feedback signal for
user-vector updates is instantiated with the keyword-based
heuristic described in Section~\ref{sec:method-rl}, requiring no
additional LLM call.

\subsection{System Modes and Baselines}
\label{sec:setup-modes}

We compare six system modes under the same frozen-backbone setting (rows in Table~\ref{tab:msc-main}): \textbf{Vanilla} (no memory), \textbf{Contextual} (full history appended), \textbf{All-memory} (all extracted preferences appended), \textbf{Reflection} (session-level reflection summaries appended to future prompts), \textbf{RAG} (dense retrieval + reranking, no user vector), and \textbf{VARS} (our full method with learned user state). Full descriptions are provided in Appendix~\ref{app:system-modes}. Global preference injection (Section~\ref{sec:method-memory}) applies identically to all modes that use preference extraction (All-memory, RAG, VARS); Vanilla and Contextual have no preference memory, and Reflection uses its own session-level summaries. All modes are evaluated on the same $60$ profiles over $60$ sessions.

\subsection{Evaluation Metrics}
\label{sec:setup-metrics}

The main comparison (Table~\ref{tab:msc-main}) reports \textbf{success rate}, \textbf{timeout rate} (fraction of sessions exhausting all turns without task completion), and \textbf{user effort} (average user tokens per session; note that \citet{mehri2026multisessioncollab} define user effort as the number of preference enforcement instances---we adopt a token-based measure as a more direct proxy for interaction cost), distinguishing whether the task is completed from how much user-side intervention is required. Table~\ref{tab:extra-metrics} adds efficiency and compliance metrics including interaction efficiency (successes per 1k user tokens), late-session success, quick resolution, first-turn enforcement, and zero-enforcement success.

\section{Result}

\label{sec:results}

\begin{figure*}[t]
  \centering
  \includegraphics[width=0.88\textwidth]{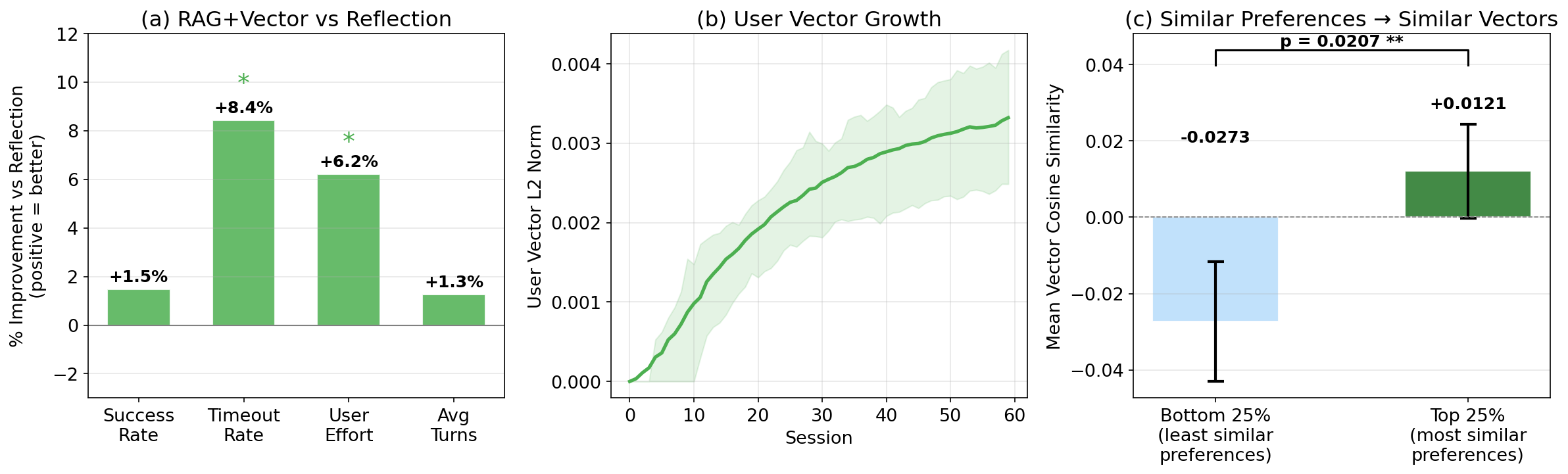}
  \caption{Main results summary on \textsc{MultiSessionCollab}.}
  \label{fig:main-results}
\end{figure*}

\subsection{Main Comparison}
\label{sec:results-main}

% Table~\ref{tab:msc-main} summarizes the main results across
% all three task domains (math-hard, math-500, bigcodebench).
% We report task success, timeout rate, and average user tokens
% per session for the six system modes
% (Section~\ref{sec:setup-modes}), evaluated over
% $60$~profiles~$\times$~$60$~sessions ($3{,}600$ sessions per
% method).

% During evaluation we identified a small fraction of sessions
% exhibiting an agent repetition bug, in which the agent produces
% near-identical responses across consecutive turns despite user
% corrections (detected via SequenceMatcher ratio $> 0.85$ on
% $\geq 2$ adjacent response pairs).
% These sessions affected $7.1\%$ of VARS sessions and
% $3.8\%$ of Reflection sessions; all numbers reported below
% exclude these sessions.

Table~\ref{tab:msc-main} summarizes results across all three
task domains for the six system modes, evaluated over
$60$~profiles~$\times$~$60$~sessions ($3{,}600$ sessions per
method).

\begin{table}[t]
  \centering
  \small
  \resizebox{\columnwidth}{!}{%
  \begin{tabular}{lccc}
    \toprule
    Method
    & Success (\%) $\uparrow$
    & Timeout (\%) $\downarrow$
    & User tokens $\downarrow$ \\
    \midrule
    VARS        & \underline{\textbf{55.2}} & \underline{\textbf{26.4}} & \underline{193.6} \\
    Reflection        & \underline{54.4} & \underline{28.8} & 207.5 \\
    Vanilla           & 54.3 & 29.2 & 232.9 \\
    Contextual        & 52.4 & 31.4 & 213.7 \\
    RAG               & 52.0 & 44.3 & \underline{\textbf{188.4}} \\
    All-memory        & 50.9 & 33.4 & 226.8 \\
    \bottomrule
  \end{tabular}}
  \caption{Main results on \textsc{MultiSessionCollab} ($60$~profiles~$\times$~$60$~sessions). Bold = best; underline = second best.}
  \label{tab:msc-main}
\end{table}

% \paragraph{Overall ranking.}
% VARS achieves the highest task success rate ($55.2\%$)
% among all six methods, followed closely by Reflection ($54.4\%$)
% and Vanilla ($53.4\%$).
% VARS also attains the lowest timeout rate ($26.4\%$),
% meaning fewer sessions exhaust all available turns without
% task completion.
% In terms of user effort, VARS requires $193.6$ user
% tokens per session on average, a $6.7\%$ reduction compared to
% Reflection ($207.5$) and a $16.9\%$ reduction compared to
% Vanilla ($232.9$).

% The gap between RAG ($52.0\%$) and VARS ($55.2\%$)
% isolates the contribution of the learned user vector:
% the $+3.2$ percentage-point improvement comes entirely from
% user-vector--modulated retrieval, since all other components
% (preference extraction, embedding model, reranker) are
% identical.

\paragraph{Overall ranking.}
VARS leads on all three primary metrics (Table~\ref{tab:msc-main}). The $+3.2$\,pp success gap over RAG isolates the user vector's contribution, since all other pipeline components are shared.

\paragraph{Comparison with Reflection.}
Paired tests across $60$ profiles (Table~\ref{tab:significance}) show VARS significantly reduces timeout ($-2.4$\,pp, $p = 0.046$) and user effort ($-13.9$ tokens, $p = 0.021$), while the success difference ($+0.9$\,pp) is not significant ($p = 0.276$).

\begin{table}[t]
  \centering
  \small
  \resizebox{\columnwidth}{!}{%
  \begin{tabular}{lcccc}
    \toprule
    Metric
    & Reflection
    & RAG+Vec
    & $\Delta$
    & $p$ (paired $t$) \\
    \midrule
    Success (\%)     & 54.4 & 55.2 & $+$0.9  & 0.276 \\
    Timeout (\%)     & 28.8 & 26.4 & $-$2.4  & \textbf{0.046$^*$} \\
    User tokens      & 207.5 & 193.6 & $-$13.9 & \textbf{0.021$^*$} \\
    \bottomrule
  \end{tabular}}
  \caption{VARS vs.\ Reflection ($60$ profiles, one-sided paired $t$-tests). $^*$\,$p < 0.05$.}
  \label{tab:significance}
\end{table}

\paragraph{Additional metrics.}
Table~\ref{tab:extra-metrics} reports efficiency and compliance metrics. VARS leads on all reported measures, yielding the highest interaction efficiency ($2.83$ successes per $1{,}000$ user tokens, $+8.4\%$ over Reflection). VARS has a slightly higher per-turn enforcement rate ($0.186$ vs.\ $0.175$, n.s.), but this does not translate into worse interaction quality given matched success with fewer turns and lower token cost.

\begin{table}[t]
  \centering
  \small
  \resizebox{\columnwidth}{!}{%
  \begin{tabular}{lccc}
    \toprule
    Metric & Reflection & RAG & VARS \\
    \midrule
    Succ.\ / 1k user tokens $\uparrow$
      & 2.61 & 2.80 & \textbf{2.83} \\
    Late success (sess.\ 30--59) $\uparrow$
      & 51.8\% & 51.7\% & \textbf{54.1\%} \\
    Quick resolution ($\leq$4 turns) $\uparrow$
      & 3.5\% & 3.4\% & \textbf{3.9\%} \\
    Zero-enf.\ success $\uparrow$
      & 60.2\% & 57.1\% & \textbf{60.6\%} \\
    First-turn enforced $\downarrow$
      & 7.9\% & 7.2\% & \textbf{7.1\%} \\
    Profiles improved $\uparrow$
      & 31.6\% & --- & \textbf{36.7\%} \\
    \bottomrule
  \end{tabular}}

  \caption{Additional metrics. $\uparrow$/$\downarrow$ = higher/lower is better. Bold = best.}
  \label{tab:extra-metrics}
\end{table}

VARS also maintains the highest late-session success and the lowest first-turn enforcement rate, indicating that cross-session learning and proactive preference surfacing both improve over time.

%% ----------------------------------------------------------------
\subsection{User-Vector Representation Analysis}
\label{sec:results-user}

We next ask whether the learned long-term user vectors
$z^{(L)}_U$ capture meaningful structure in the underlying
preference space.

% \paragraph{Setup.}
% For each pair of users $(u, v)$ we compute
% (i)~the Jaccard similarity of their \emph{revealed}
% preferences (i.e., preferences that were explicitly enforced
% at least once during the $60$ sessions), and
% (ii)~the cosine similarity of their learned long-term vectors
% $\cos(z^{(L)}_U, z^{(L)}_v)$.
% All $60$ users develop non-zero long-term vectors over the
% course of training; the mean $\ell_2$ norm grows monotonically
% from $0$ (session~$0$) to $0.0033$ (session~$60$).

\paragraph{Setup.}
For each user pair $(u, v)$ we compute the Jaccard similarity of
their revealed preferences (enforced at least once during the
$60$ sessions) and the cosine similarity of their learned
long-term vectors $\cos(z^{(L)}_U, z^{(L)}_v)$. All $60$ users
develop non-zero long-term vectors; the mean $\ell_2$ norm grows
monotonically from $0$ (session~$0$) to $0.0033$ (session~$60$).

% \paragraph{Preference overlap predicts vector similarity.}
% We test whether users with more similar revealed preferences
% end up with more similar long-term vectors.
% A Spearman rank correlation between Jaccard overlap and
% vector cosine similarity yields $\rho = 0.040$
% ($p = 0.093$).
% While the global correlation is modest, a quartile analysis
% reveals a clearer signal: user pairs in the top~$25\%$ of
% preference similarity have a mean vector cosine of $+0.012$,
% whereas pairs in the bottom~$25\%$ have a mean cosine of
% $-0.027$ (Mann--Whitney $U$ test, $p = 0.021$).
% This confirms that users who share more preferences are
% pushed toward more similar regions in vector space by the
% REINFORCE updates, consistent with the collaborative-filtering
% intuition described in Section~\ref{sec:method-rl}.

\paragraph{Preference overlap predicts vector similarity.}
Spearman correlation between Jaccard overlap and vector cosine is
$\rho = 0.040$ ($p = 0.093$)—modest globally, but a quartile
analysis reveals a clearer signal: user pairs in the top~$25\%$
of preference similarity have mean vector cosine $+0.012$ versus
$-0.027$ for the bottom~$25\%$ (Mann--Whitney $U$,
$p = 0.021$), confirming that shared preferences push users
toward similar regions in vector space via REINFORCE updates.

% \paragraph{Dual-vector separation.}
% Table~\ref{tab:vector-analysis} compares the three vector
% variants.
% The long-term vector $z^{(L)}$ shows a significant
% association with long-term preference overlap (quintile test
% $p = 0.006$), while the short-term vector $z^{(S)}$ does not
% ($p = 0.586$).
% This is the expected behavior: $z^{(L)}$ accumulates stable
% preference information across sessions, whereas $z^{(S)}$ is
% designed to capture transient, within-session context and
% decays between sessions.
% The fact that $z^{(S)}$ does \emph{not} correlate with
% long-term preference structure confirms that the dual-vector
% design successfully separates stable user identity from
% session-specific adaptation.

\paragraph{Dual-vector separation.}
Table~\ref{tab:vector-analysis} compares the three vector
variants. The long-term vector $z^{(L)}$ significantly associates
with preference overlap (quintile $p = 0.006$), while $z^{(S)}$
does not ($p = 0.586$), confirming that the dual-vector design
separates stable user identity from session-specific adaptation.

\begin{table}[t]
  \centering
  \small
  \resizebox{\columnwidth}{!}{%
  \begin{tabular}{lcc}
    \toprule
    Vector & Spearman $\rho$ ($p$) & Quintile $p$ \\
    \midrule
    $z^{(L)}$ (long-term)  & 0.040\;(0.093) & \textbf{0.006} \\
    $z^{(S)}$ (short-term) & 0.010\;(0.681) & 0.586 \\
    Combined               & 0.046\;(0.054) & 0.003 \\
    \bottomrule
  \end{tabular}}

  \caption{
    User-vector cosine vs.\ preference Jaccard overlap ($\binom{60}{2}$ pairs).}
  \label{tab:vector-analysis}
\end{table}

%% ----------------------------------------------------------------
\paragraph{Dual-vector ablation.}
\label{sec:results-ablation}

Table~\ref{tab:ablation} reports ablations that disable one or both user vectors. The full dual-vector model leads on all primary metrics. Removing $z^{(L)}$ hurts more than removing $z^{(S)}$ in terms of timeout rate, while removing $z^{(S)}$ has a larger effect on success rate. The non-timeout user token column reveals a functional separation: $z^{(L)}$ drives effort reduction in successful sessions (Full $166.5$ $\approx$ $z^{(L)}$-only $168.1$, both well below $z^{(S)}$-only $185.7$), while $z^{(S)}$ contributes more to timeout avoidance ($z^{(S)}$-only $12.7\%$ vs.\ $z^{(L)}$-only $14.7\%$). This supports the intended design: $z^{(L)}$ captures stable preferences that reduce corrective interaction, and $z^{(S)}$ enables within-session adaptation that prevents session failure.

\begin{table}[t]
  \centering
  \small
  \resizebox{\columnwidth}{!}{%
  \begin{tabular}{lcccc}
    \toprule
    Method
    & Success (\%) $\uparrow$
    & Timeout (\%) $\downarrow$
    & User tok $\downarrow$
    & Non-TO utok $\downarrow$ \\
    \midrule
    VARS (full) & \textbf{74.0} & \textbf{10.7} & \textbf{474.5} & \textbf{166.5} \\
    $z^{(S)}$ only    & 72.7 & 12.7 & 846.8 & 185.7 \\
    $z^{(L)}$ only    & 71.3 & 14.7 & 1347.4 & 168.1 \\
    No vector         & 70.0 & 14.0 & 1566.1 & 183.7 \\
    \bottomrule
  \end{tabular}}

  \caption{
    Dual-vector ablation on math-hard ($10$~profiles~$\times$~$15$~sessions) with GPT-4o-mini. Higher absolute success reflects the stronger backbone. Non-TO utok = non-timeout sessions only.}
  \label{tab:ablation}
\end{table}

We also identify three recurring failure modes—over-cautious clarification, preference overload, and early user disengagement—which are more pronounced for RAG ($44.3\%$ timeout) than VARS ($26.4\%$); details are in Appendix~\ref{app:failure-modes}.

\section{Discussion}

\label{sec:discussion}

\paragraph{Where does personalization help?}
The user vector's primary contribution is not raw success rate—where
the gain over Reflection is modest and not significant—but
interaction efficiency: matching task success with significantly less
user effort and fewer timeouts. This suggests that under frozen
backbones, lightweight user modeling improves \emph{how} the agent
interacts rather than \emph{how often} it ultimately succeeds.
Comparing RAG ($52.0\%$) with VARS ($55.2\%$) further shows
that the vector mitigates preference overload: without it, retrieval
surfaces an increasingly noisy set of preferences as memory grows,
leading the agent to hedge or produce unfocused responses (RAG
timeout: $44.3\%$ vs.\ VARS: $26.4\%$). That Vanilla
($54.3\%$) nearly matches Reflection ($54.4\%$) despite having no
cross-session memory reinforces this point—a meaningful portion of
success is driven by intrinsic problem-solving ability, making
interaction efficiency a more informative measure of personalization
quality than success rate alone. We note that aggressive preference surfacing can risk over-personalization \citep{hu2026opbench}; our design mitigates this through selective retrieval biasing and a cap on global preference injection, though systematic evaluation of this risk remains future work.

\paragraph{Preference format and agent compliance.}
Directly injecting structured condition--action rules into an 8B-parameter agent's prompt often fails to elicit compliance: the agent treats them as metadata rather than instructions. Reflection's natural-language summaries are more readily followed, and the \textsc{MultiSessionCollab} prompt template is designed for reflection-style notes, suggesting that compliance is bottlenecked by presentation format as well as retrieval quality.

\section{Conclusion}

\label{sec:conclusion}
We have presented a frozen-backbone personalization framework that
represents each user as a learned dual vector in a shared preference
space, updated online from weak scalar rewards and used to modulate
retrieval over structured preference memory---without modifying any backbone model. This lightweight approach matches a reasoning-based baseline in task success while
significantly reducing user effort and timeout rate, and the learned
vectors encode interpretable preference structure that separates
stable user identity from session-specific context. These results motivate scaling to richer preference types, stronger
reward signals, and real-user evaluation.

\section*{Limitations}

\label{sec:limitations}

\paragraph{Scale and generalization.}
We evaluate on $60$ profiles drawn from a pool of $200$, with $60$ sessions per profile. While this yields $3{,}600$ sessions per method, it remains a single benchmark with LLM-simulated users. Extending to real users, longer interaction horizons, and non-style preferences (e.g., topical interests, tool choices, safety constraints) is necessary to assess generalizability.

\paragraph{Reward signal.}
The current system uses a keyword-based heuristic to classify user follow-ups into reward categories. While fast and requiring no additional model, this heuristic may miss subtle feedback cues. We have implemented and validated an LLM-as-judge alternative (\texttt{Llama-3.1-8B-Instruct}, $83$--$92\%$ agreement with GPT-based judges) that can serve as a drop-in replacement for richer reward estimation, but have not yet evaluated its effect on user-vector learning dynamics.

\paragraph{Hyperparameter sensitivity.}
Learning rates ($\eta_L$, $\eta_S$), scoring weights ($\beta_L$, $\beta_S$), decay rate ($\lambda$), and retrieval parameters (top-$k$, reranker threshold) are set heuristically (Appendix~\ref{app:hyperparams}). Systematic hyperparameter sweeps are needed to understand sensitivity.

\paragraph{Simulator fidelity.}
All evaluation relies on an LLM-based user simulator whose preference enforcement behavior may differ from real users. The simulator's tolerance thresholds and enforcement patterns introduce variance that may not reflect authentic human interaction dynamics \citep{mehri2025goal}. More broadly, LLM-based user simulators can exhibit systematic biases---such as unrealistic patience or overly formulaic preference enforcement---that may inflate or deflate measured personalization gains relative to real users.

\paragraph{Privacy.}
Storing persistent user vectors and structured preference memories introduces profiling risks. In deployment, these artifacts should be subject to the same access controls and deletion policies as conversation history.

\bibliography{custom}

\appendix

\section{Preference Extractor Training Details}
\label{app:extractor}

\paragraph{Model and objective.}
The preference extractor $M_{\text{ext}}$ is a 0.6B-parameter
Qwen3 model~\cite{yang2025qwen3} finetuned for
instruction-following and structured JSON output with
LLaMA-Factory~\cite{zheng2024llamafactory}. Given a short dialogue
window $W_t$ (typically the last few turns up to $t$),
$M_{\text{ext}}$ is trained to predict a set of preference tuples
\newline $P_t = \{(\text{condition}_j, \text{action}_j)\}_{j=1}^{J_t}$ in a
fixed JSON schema (Section~\ref{sec:method-extractor}). Each tuple
describes when a preference applies (condition) and what the agent
should do (action). The model is trained with a standard
left-to-right language modeling objective on the JSON-serialized
output.

\paragraph{Data construction.}
We construct a mixed dataset of $564{,}423$ examples
($33\%$ positive, $67\%$ negative) from the sources listed in
Table~\ref{tab:extractor-data}. Teacher labels are generated by
GPT-5.1 via the OpenAI Batch API; we retain both positive
examples (with extracted preferences) and negative examples
(no preference) so that $M_{\text{ext}}$ learns when
\emph{not} to emit any preference. All sources are converted
into a unified instruction format with the dialogue window
$W_t$ as input and the target JSON as output, then randomly
mixed during training.

\begin{table}[t]
  \centering
  \small
  \resizebox{\columnwidth}{!}{%
  \begin{tabular}{lr}
    \toprule
    Source & Count \\
    \midrule
    GPT-5.1 synthesized positives  & 156{,}489 \\
    LMSYS-Chat-1M~\cite{zheng2023lmsyschat1m}       & 133{,}086 \\
    WildChat~\cite{zhao2024wildchat}           & 108{,}677 \\
    Retry / recovery labels        &  68{,}778 \\
    Alpaca-Cleaned~\cite{wang2023camels}     &  42{,}013 \\
    ShareGPT-Vicuna\footnotemark &  33{,}025 \\
    SlimOrca~\cite{lian2023slimorca} &  22{,}345 \\
    Manual fixes                   &      10 \\
    \midrule
    \textbf{Total}                 & \textbf{564{,}423} \\
    \bottomrule
  \end{tabular}}
  \caption{Training data sources for the preference extractor.}
  \label{tab:extractor-data}
\end{table}
\footnotetext{\url{https://huggingface.co/datasets/anon8231489123/ShareGPT_Vicuna_unfiltered}}

\paragraph{Training hyperparameters.}
We perform full supervised finetuning (SFT) with a global batch
size of 128, AdamW optimizer, learning rate
$2\times 10^{-5}$, a cosine learning rate schedule, and
bf16 precision on H200 GPUs. We train for a small number of epochs
until validation loss plateaus. On a held-out evaluation set of teacher-labeled examples, the
resulting model achieves $99.7\%$ JSON validity, $97.5\%$ recall,
$37.7\%$ precision, and $54.4\%$ F1. The high recall ensures that
nearly all explicitly stated preferences are captured, while the
lower precision reflects a deliberate ``extract first, filter later''
design: over-extracted preferences are filtered downstream by the
reranker and user-vector scoring (Section~\ref{sec:method-user-vector}).

\section{User Profile Example}
\label{app:profiles}

This appendix provides a concrete user profile from our
\textsc{MultiSessionCollab} math-hard experiments. The profile
(\texttt{user\_14b429db}) is one of the five users used in the main
experiments (Section~\ref{sec:setup-profiles}); it encodes 43 structured
style preferences and 15 conflict groups, and was run for 20 sessions
in our pilot study. We also show the subset of preferences that were
explicitly revealed during the 20-session interaction, and the basic
interaction statistics for this user.

\subsection{Full Preference Schema for \texttt{user\_14b429db}}

\begin{lstlisting}[style=jsonprofile]
{
  "user_id": "user_14b429db",
  "persona": "A senior backend engineer who values efficiency and directness. Prefers practical solutions over theoretical discussions.",
  "meta": {
    "total_preferences": 43,
    "total_conflict_groups": 15,
    "generator": "schema_based"
  },
  "preferences": [
    {
      "id": "cs_008",
      "condition": "providing error handling examples",
      "action": "always use specific exception types, never bare except",
      "conflict_group": null,
      "priority_context": ["error", "exception", "try"]
    },
    {
      "id": "rf_003",
      "condition": "providing a final answer or conclusion",
      "action": "put the answer first, then explanation",
      "conflict_group": "answer_position",
      "priority_context": ["direct_question", "what_is"]
    },
    {
      "id": "ds_001",
      "condition": "discussing machine learning concepts",
      "action": "include the mathematical formulation alongside intuitive explanation",
      "conflict_group": null,
      "priority_context": ["ml", "machine learning", "model"]
    },
    {
      "id": "ds_006",
      "condition": "writing or reviewing documentation",
      "action": "be concise, avoid marketing language, focus on usage",
      "conflict_group": null,
      "priority_context": ["documentation", "docs", "readme"]
    },
    {
      "id": "vb_004",
      "condition": "I'm debugging and say 'it doesn't work'",
      "action": "focus on diagnosing the specific issue, skip general explanations",
      "conflict_group": "explanation_depth",
      "priority_context": ["debugging", "error", "fix"]
    },
    {
      "id": "ip_002",
      "condition": "I give a clear and specific instruction",
      "action": "execute directly without asking for confirmation",
      "conflict_group": "autonomy",
      "priority_context": ["do this", "make this", "specific_instruction"]
    },
    {
      "id": "vb_002",
      "condition": "the topic involves complex algorithms or mathematics",
      "action": "provide detailed step-by-step derivation with intermediate results",
      "conflict_group": "response_length",
      "priority_context": ["complex_topic", "proof", "derivation"]
    },
    {
      "id": "ec_001",
      "condition": "I make a minor error in terminology",
      "action": "correct it gently inline without making it a focus",
      "conflict_group": "correction_style",
      "priority_context": ["minor_error", "terminology"]
    },
    {
      "id": "ms_002",
      "condition": "discussing statistics or probability",
      "action": "start with intuition and real-world interpretation before formulas",
      "conflict_group": "math_approach",
      "priority_context": ["probability", "statistics", "likelihood"]
    },
    {
      "id": "ip_001",
      "condition": "the task is complex with multiple parts",
      "action": "confirm the plan before executing, break into phases",
      "conflict_group": "autonomy",
      "priority_context": ["complex", "multiple", "project"]
    },
    {
      "id": "cs_004",
      "condition": "the code snippet is short (under 20 lines)",
      "action": "include inline comments explaining each logical block",
      "conflict_group": "comment_style",
      "priority_context": ["example", "snippet"]
    },
    {
      "id": "vb_001",
      "condition": "I say 'quick question' or 'briefly'",
      "action": "respond in 3 sentences or fewer, no elaboration",
      "conflict_group": "response_length",
      "priority_context": ["time_pressure", "simple_query"]
    },
    {
      "id": "rf_002",
      "condition": "explaining a sequential process or procedure",
      "action": "use numbered steps with clear transitions",
      "conflict_group": "format_structure",
      "priority_context": ["tutorial", "how-to", "setup"]
    },
    {
      "id": "ms_005",
      "condition": "discussing proofs",
      "action": "structure as: claim, approach sketch, formal proof, intuition recap",
      "conflict_group": "proof_style",
      "priority_context": ["prove", "proof", "show that"]
    },
    {
      "id": "ec_004",
      "condition": "I ask you to correct your previous response",
      "action": "acknowledge the error explicitly, then provide corrected version",
      "conflict_group": null,
      "priority_context": ["you were wrong", "that's not right", "actually"]
    },
    {
      "id": "ec_003",
      "condition": "my code has a bug",
      "action": "show the bug location, explain why it's wrong, provide the fix",
      "conflict_group": null,
      "priority_context": ["bug", "error", "wrong"]
    },
    {
      "id": "cs_001",
      "condition": "writing Python code",
      "action": "use snake_case for variables and functions, include type hints",
      "conflict_group": "naming_convention",
      "priority_context": ["python", "py"]
    },
    {
      "id": "oa_003",
      "condition": "any response with code",
      "action": "always specify the language in the code fence",
      "conflict_group": null,
      "priority_context": ["code"]
    },
    {
      "id": "ip_006",
      "condition": "I thank you or say the answer was helpful",
      "action": "don't add unnecessary follow-up, just acknowledge briefly",
      "conflict_group": null,
      "priority_context": ["thanks", "helpful", "great"]
    },
    {
      "id": "rf_004",
      "condition": "teaching a new concept",
      "action": "build up intuition before giving the formal definition",
      "conflict_group": "answer_position",
      "priority_context": ["learning", "explain", "why"]
    },
    {
      "id": "ds_004",
      "condition": "explaining a theoretical concept",
      "action": "start with definition, then example, then edge cases",
      "conflict_group": "example_position",
      "priority_context": ["concept", "theory", "what is"]
    },
    {
      "id": "vb_005",
      "condition": "I explicitly share my current understanding first",
      "action": "acknowledge what I got right, then correct only the gaps",
      "conflict_group": null,
      "priority_context": ["validation", "checking"]
    },
    {
      "id": "cs_006",
      "condition": "I ask for a code review",
      "action": "focus only on bugs and logic errors, ignore style issues",
      "conflict_group": "review_scope",
      "priority_context": ["review", "check", "look at"]
    },
    {
      "id": "cs_003",
      "condition": "writing SQL queries",
      "action": "use UPPERCASE for keywords, lowercase for table/column names",
      "conflict_group": "naming_convention",
      "priority_context": ["sql", "database", "query"]
    },
    {
      "id": "ec_002",
      "condition": "I have a fundamental misconception",
      "action": "address the misconception directly and clearly before proceeding",
      "conflict_group": "correction_style",
      "priority_context": ["misconception", "fundamental_error"]
    },
    {
      "id": "ms_003",
      "condition": "I ask to verify my calculation",
      "action": "check my work step by step, point out where I diverged if wrong",
      "conflict_group": null,
      "priority_context": ["verify", "check", "is this right"]
    },
    {
      "id": "ms_001",
      "condition": "solving algebraic equations",
      "action": "show each manipulation step with the operation applied noted",
      "conflict_group": "math_detail",
      "priority_context": ["solve", "equation", "algebra"]
    },
    {
      "id": "vb_003",
      "condition": "I ask 'why' or 'how come'",
      "action": "always explain the underlying reasoning, not just the what",
      "conflict_group": "explanation_depth",
      "priority_context": ["curiosity", "understanding"]
    },
    {
      "id": "cs_002",
      "condition": "writing JavaScript or TypeScript code",
      "action": "use camelCase for variables, PascalCase for classes",
      "conflict_group": "naming_convention",
      "priority_context": ["javascript", "js", "typescript", "ts"]
    },
    {
      "id": "ds_003",
      "condition": "discussing APIs or library usage",
      "action": "show a minimal working example before explaining parameters",
      "conflict_group": "example_position",
      "priority_context": ["api", "library", "how to use"]
    },
    {
      "id": "oa_004",
      "condition": "providing commands to run",
      "action": "use bash code blocks, include expected output as comments",
      "conflict_group": null,
      "priority_context": ["command", "run", "terminal"]
    },
    {
      "id": "ms_004",
      "condition": "the problem involves calculus",
      "action": "state the rule being applied (chain rule, integration by parts, etc.)",
      "conflict_group": "math_detail",
      "priority_context": ["derivative", "integral", "calculus"]
    },
    {
      "id": "ms_006",
      "condition": "I'm practicing for an exam",
      "action": "after solving, give a similar practice problem",
      "conflict_group": null,
      "priority_context": ["practice", "exam", "test"]
    },
    {
      "id": "ds_002",
      "condition": "discussing system design or architecture",
      "action": "describe components as a list first, then explain interactions",
      "conflict_group": null,
      "priority_context": ["design", "architecture", "system"]
    },
    {
      "id": "ip_003",
      "condition": "I seem uncertain or ask 'what do you think'",
      "action": "provide a recommendation with brief rationale, not just options",
      "conflict_group": "guidance_style",
      "priority_context": ["uncertain", "should I", "what do you think"]
    },
    {
      "id": "ds_005",
      "condition": "discussing data structures",
      "action": "always include time complexity for operations mentioned",
      "conflict_group": null,
      "priority_context": ["data structure", "array", "tree", "hash"]
    },
    {
      "id": "ip_004",
      "condition": "I'm comparing alternatives",
      "action": "present trade-offs in a table format with clear criteria",
      "conflict_group": "guidance_style",
      "priority_context": ["compare", "vs", "or", "which"]
    },
    {
      "id": "ip_005",
      "condition": "I express frustration or say 'this is annoying'",
      "action": "acknowledge the difficulty briefly, then provide direct help",
      "conflict_group": null,
      "priority_context": ["frustration", "annoying", "stuck"]
    },
    {
      "id": "oa_001",
      "condition": "generating code that will be copied",
      "action": "provide code in a single copyable block, no interleaved explanation",
      "conflict_group": "code_presentation",
      "priority_context": ["copy", "use this", "give me code"]
    },
    {
      "id": "oa_002",
      "condition": "teaching through code examples",
      "action": "break code into small chunks with explanation between each",
      "conflict_group": "code_presentation",
      "priority_context": ["teach", "learn", "understand"]
    },
    {
      "id": "cs_005",
      "condition": "the code is a complete module or class",
      "action": "use docstrings at function/class level, minimal inline comments",
      "conflict_group": "comment_style",
      "priority_context": ["module", "class", "production"]
    },
    {
      "id": "rf_001",
      "condition": "listing multiple items or options",
      "action": "use bullet points with consistent indentation",
      "conflict_group": "format_structure",
      "priority_context": ["enumeration", "comparison"]
    },
    {
      "id": "cs_007",
      "condition": "I ask to improve or refactor code",
      "action": "address both logic and style, suggest modern idioms",
      "conflict_group": "review_scope",
      "priority_context": ["improve", "refactor", "better"]
    }
  ],
  "conflict_groups": {
    "answer_position": ["rf_003", "rf_004"],
    "explanation_depth": ["vb_004", "vb_003"],
    "autonomy": ["ip_002", "ip_001"],
    "response_length": ["vb_002", "vb_001"],
    "correction_style": ["ec_001", "ec_002"],
    "math_approach": ["ms_002"],
    "comment_style": ["cs_004", "cs_005"],
    "format_structure": ["rf_002", "rf_001"],
    "proof_style": ["ms_005"],
    "naming_convention": ["cs_001", "cs_003", "cs_002"],
    "example_position": ["ds_004", "ds_003"],
    "review_scope": ["cs_006", "cs_007"],
    "math_detail": ["ms_001", "ms_004"],
    "guidance_style": ["ip_003", "ip_004"],
    "code_presentation": ["oa_001", "oa_002"]
  }
}
\end{lstlisting}

\subsection{Revealed Preferences and Per-User Metrics}

During the 20-session run used in our main experiments,
the user simulator explicitly revealed a subset of the
ground-truth preferences for \texttt{user\_14b429db}. The
table below summarizes the 10 revealed preferences for this
user.

\begin{table*}[t]
  \centering
  \begin{tabularx}{\textwidth}{lXX}
    \toprule
    ID & When & Then \\
    \midrule
    cs\_008 & providing error handling examples &
      always use specific exception types, never bare except \\
    ds\_001 & discussing ML concepts &
      include mathematical formulation alongside intuitive explanation \\
    ds\_006 & writing documentation &
      be concise, avoid marketing language, focus on usage \\
    vb\_002 & complex algorithms/mathematics &
      provide detailed step-by-step derivation with intermediate results \\
    ms\_002 & discussing statistics/probability &
      start with intuition and real-world interpretation before formulas \\
    rf\_003 & providing a final answer &
      put the answer first, then explanation \\
    ip\_002 & clear and specific instruction &
      execute directly without asking for confirmation \\
    vb\_004 & debugging ``it doesn't work'' &
      focus on diagnosing specific issue, skip general explanations \\
    ip\_001 & complex multi-part task &
      confirm plan before executing, break into phases \\
    ec\_001 & minor terminology error &
      correct gently inline without making it a focus \\
    \bottomrule
  \end{tabularx}
  \caption{Revealed preferences for \texttt{user\_14b429db}
  during the 20-session MultiSessionCollab run.}
  \label{tab:profile-revealed-14b4}
\end{table*}

In this particular run, \texttt{user\_14b429db} had 20 sessions,
with a task success rate of 65\% (13/20 successful conversations)
and a total of 48 explicit enforcement turns from the user
simulator.

\section{Offline Style-Persona Experiments}
\label{app:offline}

Before the MultiSessionCollab benchmark was released in Jan.6.2026, we conducted a
set of offline style-persona experiments in a simplified,
rule-based simulation environment. The goal was to isolate the
effects of preference memory and online user vectors on cross-session
style adherence, using easy tasks and discrete style preferences.

\subsection{Environment and personas}

The environment consists of synthetic users endowed with latent
style preferences over four dimensions:
response length, bullet usage, and language
(English vs.\ Chinese). Each user is assigned to one of several
personas such as
A\_short\_bullets\_en,
B\_short\_no\_bullets\_en,
C\_long\_bullets\_en,
D\_short\_bullets\_zh, and
E\_long\_no\_bullets\_zh, which specify a ground-truth
\texttt{StylePrefs} tuple
$(\texttt{require\_short}, \texttt{max\_chars},
  \texttt{require\_bullets}, \texttt{lang})$.

For each persona we script multiple sessions with three phases:
(i) a preference-reveal session where the user explicitly states
length, bullet, and language preferences; (ii) a cross-session
retention session where the user only issues tasks without
restating preferences; and (iii) mixed sessions where the user
sometimes restates preferences or complains when they are violated,
providing additional feedback. Tasks are simple list and QA
prompts (e.g., ``List three healthy breakfast ideas.'’, ‘‘What is the
capital of France?’’) instantiated in the persona’s preferred
language.

We consider three system modes: VANILLA (no preference memory),
STATIC-MEM (preference memory without user vectors), and
ONLINE-USER (preference memory plus online user vectors). All
share the same frozen chat, embedding, and reranker models; only
the external memory and user vectors differ.

\subsection{Metrics}

We evaluate personalization with the same style-oriented metrics
as in the main text: turn-level satisfaction scores, violation
rates for different error types (e.g., too\_long, no\_bullets,
wrong\_lang), and Recall@k of relevant preference memories.
Formally, for Session~2 we compute
\[
  \text{AvgSatS2}
    = \frac{1}{|T^{\text{base}}_{\text{S2}}|}
      \sum_{t\in T^{\text{base}}_{\text{S2}}} s_t,
\]
and for each violation type $v$
\[
  \text{ViolRateS2}(v)
    = \frac{\#\{t \in T^{\text{base}}_{\text{S2}} : v \in \text{Viol}_t\}}
           {|T^{\text{base}}_{\text{S2}}|},
\]
where $s_t$ is the rule-based satisfaction score and
$\text{Viol}_t$ are violation labels at turn $t$. Memory
Recall@k is defined as the fraction of turns where at least one
selected memory card encodes the relevant preference.

\subsection{Results with three-session lifetimes}

Table~\ref{tab:offline-3session} reports Session~2 results for
ONLINE-USER and the VANILLA baseline under a strict retention
setting where complaint turns are disabled. ONLINE-USER achieves
substantially higher satisfaction and lower violation rates, and
retrieves relevant preference memories almost perfectly.

\begin{table}[t]
  \centering
  \resizebox{\columnwidth}{!}{%
  \begin{tabular}{l S[table-format=1.4] S[table-format=1.4]}
    \toprule
    Metric & {ONLINE-USER} & {VANILLA} \\
    \midrule
    AvgSatS2 & 0.9500 & 0.7250 \\
    ViolRateS2(too\_long) & 0.1667 & 0.4167 \\
    ViolRateS2(no\_bullets) & 0.0000 & 0.5000 \\
    Recall@k\_S2(SHORT) & 0.6667 & 0.0000 \\
    Recall@k\_S2(BULLETS) & 0.8333 & 0.0000 \\
    Recall@k\_S2(LANG) & 1.0000 & 0.0000 \\
    \bottomrule
  \end{tabular}}
  \caption{Offline Session~2 results (no complaints, 6 personas).
  ONLINE-USER substantially improves satisfaction and reduces
  violations compared to a VANILLA LLM without preference memory.}
  \label{tab:offline-3session}
\end{table}

\subsection{Longer lifetimes and user-vector alignment}

We also extend the simulation to ten sessions per persona (with
complaint turns enabled) and again compare ONLINE-USER to
VANILLA. Table~\ref{tab:offline-10session} shows that the gains
persist under longer lifetimes: ONLINE-USER maintains higher
satisfaction and lower violation rates, while consistently
retrieving the relevant preference memories.:

\begin{table}[t]
  \centering
  \resizebox{\columnwidth}{!}{%
  \begin{tabular}{l S[table-format=1.4] S[table-format=1.4]}
    \toprule
    Metric & {ONLINE-USER} & {VANILLA} \\
    \midrule
    AvgSatS2 & 0.9750 & 0.8625 \\
    ViolRateS2(too\_long) & 0.0833 & 0.1667 \\
    ViolRateS2(no\_bullets) & 0.0000 & 0.2917 \\
    Recall@k\_S2(SHORT) & 0.6667 & 0.0000 \\
    Recall@k\_S2(BULLETS) & 0.7917 & 0.0000 \\
    Recall@k\_S2(LANG) & 1.0000 & 0.0000 \\
    \bottomrule
  \end{tabular}}
  \caption{Offline Session~2 results with longer lifetimes
  (10 sessions per user). ONLINE-USER continues to outperform
  VANILLA in satisfaction, violation rates, and preference-memory
  recall.}
  \label{tab:offline-10session}
\end{table}

Finally, we probe whether the learned user vectors capture
meaningful structure by correlating cosine similarity between
long-term user vectors with the ground-truth overlap of style
preferences across personas. In a dedicated experiment with
six synthetic personas and about 270 turns of interaction per
persona, we observe a positive Spearman correlation of
$\rho \approx 0.27$ between learned similarities and
preference-overlap similarities, despite the short histories and
highly saturated satisfaction scores.

\section{Hyperparameters and Implementation Details}
\label{app:hyperparams}

This appendix summarizes the main hyperparameters and
implementation choices for user-vector learning, preference-aware
scoring, and retrieval. All experiments in Section~\ref{sec:results}
use the \emph{actual values} listed below.

\subsection{Core REINFORCE Update Parameters}

The user-vector updates in Section~\ref{sec:method-rl} follow the
REINFORCE-style rule
\begin{align}
  \Delta \mathbf{z}^{(L)}_U
    &= \eta_L \frac{A^{\text{retr}}_t}{\tau}
      \bigl(\mathbf{v}_{\text{chosen},t} - \boldsymbol{\mu}_t\bigr),
      \nonumber\\
  \Delta \mathbf{z}^{(S)}_{U,t}
    &= \eta_S \frac{A^{\text{retr}}_t}{\tau}
      \bigl(\mathbf{v}_{\text{chosen},t} - \boldsymbol{\mu}_t\bigr),
      \nonumber
\end{align}
with short-term decay
\[
  \mathbf{z}^{(S)}_{U,t+1}
    = (1 - \lambda)\,\mathbf{z}^{(S)}_{U,t}
      + \Delta \mathbf{z}^{(S)}_{U,t},
\]
and a per-user exponential moving average baseline
\[
  b_U \leftarrow (1 - \alpha)\,b_U + \alpha\,\hat r_t.
\]

Table~\ref{tab:hyper-core} lists the corresponding hyperparameters.

\begin{table}[tb]
  \centering
  \begin{tabularx}{\columnwidth}{X}
    \toprule
    Setting \\
    \midrule
    $\eta_L = 1.0\times 10^{-2}$: LR for long-term vector $\mathbf{z}^{(L)}_U$ \\
    $\eta_S = 5.0\times 10^{-2}$: LR for short-term vector $\mathbf{z}^{(S)}_{U,t}$ \\
    $\lambda = 0.1$: Exponential decay for $\mathbf{z}^{(S)}_{U,t}$ between turns \\
    $\alpha = 0.05$: EMA coefficient for reward baseline $b_U$ \\
    $\tau = 1.0$: Policy temperature in the softmax over memories \\
    \bottomrule
  \end{tabularx}
  \caption{Core REINFORCE update parameters. All experiments
  use the code values.}
  \label{tab:hyper-core}
\end{table}

\subsection{User-Vector Weighting and Item Space}

The effective user vector used in the user-aware scoring function
(Section~\ref{sec:method-user-vector}) is
\[
  \mathbf{z}^{\text{eff}}_{U,t}
    = \beta_L\,\mathbf{z}^{(L)}_U + \beta_S\,\mathbf{z}^{(S)}_{U,t},
\]
where $\beta_L$ and $\beta_S$ control the relative weight of long-
and short-term preferences. Item vectors $\mathbf{v}_m$ are obtained
by projecting embedding vectors $\mathbf{e}_m\in\mathbb{R}^d$ into
a global $k$-dimensional PCA space (Section~\ref{sec:method-memory}).
The values used in our experiments are summarized in
Table~\ref{tab:hyper-user}.

\begin{table}[t]
  \centering

  \begin{tabularx}{\columnwidth}{X}
    \toprule
    Setting \\
    \midrule
    $\beta_L = 2.0$: Weight of long-term vector in $\mathbf{z}^{\text{eff}}_{U,t}$ \\
    $\beta_S = 5.0$: Weight of short-term vector in $\mathbf{z}^{\text{eff}}_{U,t}$ \\
    $k$ (\texttt{item\_dim}) $= 256$: Dimensionality of item and user vectors \\
    \bottomrule
  \end{tabularx}
  \caption{User-vector weighting and item-space dimension.}
  \label{tab:hyper-user}
\end{table}

\subsection{Retrieval Parameters}

Dense retrieval returns a candidate set $M_t$ of size
$K = \texttt{dense\_topk}$, and the reranker then selects the top
$\texttt{rerank\_topk}$ memories (under the user-aware score
$s(\cdot)$) to inject into the LLM prompt. The values are:

\begin{table}[t]
  \centering

  \begin{tabularx}{\columnwidth}{X}
    \toprule
    Setting \\
    \midrule
    \texttt{dense\_topk} $= 64$: Number of candidates from dense retrieval \\
    \texttt{rerank\_topk} $= 3$: Final memories fed to the LLM \\
    \bottomrule
  \end{tabularx}
  \caption{Retrieval parameters used in our RAG and VARS
  implementations.}
  \label{tab:hyper-retrieval}
\end{table}

\subsection{Reward Gating Logic}

The reward model $C$ outputs a scalar reward $\hat r_t$ and a
gating scalar $g_t\in[0,1]$ (Section~\ref{sec:method-rl}). The gate
controls how much of the reward is attributed to the retrieval policy
versus other factors (e.g., the backbone LLM or topic shifts). In
code, we implement a small set of hand-crafted gating rules based
on the reward sign and a similarity signal between the chosen
memories and the query:

\begin{table}[t]
  \centering

  \begin{tabularx}{\columnwidth}{lXc}
    \toprule
    Case & Condition & $g_t$ \\
    \midrule
    Retrieval failure
      & $\hat r_t < -0.5$, low similarity
      & $0.9$ \\
    LLM failure
      & $\hat r_t < -0.5$, high similarity
      & $0.2$ \\
    Good + retrieval helped
      & $\hat r_t > 0.5$, similarity $> 0.4$
      & $0.6$ \\
    Good + no retrieval help
      & $\hat r_t > 0.5$, similarity $\le 0.4$
      & $0.3$ \\
    Default
      & neutral / other cases
      & $0.5$ \\
    \bottomrule
  \end{tabularx}
  \caption{Gating cases for the retrieval-specific advantage
  $A^{\text{retr}}_t = g_t(\hat r_t - b_U)$. ``Similarity'' refers
  to the similarity between chosen memories and the current query;
  in the ``Good'' cases we use a numerical threshold of $0.4$.}
  \label{tab:gating}
\end{table}

These rules are intentionally simple and were chosen heuristically:
they are meant to (i) emphasize retrieval failures when the reward
is strongly negative and the retrieved memories are a poor match,
(ii) down-weight updates when failures are likely due to the LLM
rather than retrieval (negative reward but high similarity), and
(iii) give somewhat larger credit to retrieval when positive rewards
co-occur with high memory–query similarity. As discussed in
Section~\ref{sec:discussion}, a more systematic exploration of
gating and reward design is an important direction for future work.

\section{Theoretical Motivation and Exact Decomposition of the User-State Updates}
\label{app:theory}

This appendix formalizes two exact properties of the update rule in Section~3.6. First, conditional on the retrieved candidate set, the implemented updates are gradient steps on a fixed-candidate surrogate objective. Second, the dual-vector recursion admits an exact two-timescale decomposition: the long-term state accumulates all past update increments, while the short-term state is an exponentially decaying sum of recent increments. These statements are about the surrogate induced by the observed retrieved set and the heuristic attribution weight $A_t^{\mathrm{retr}}$; they are not claims that the full end-to-end retrieval-and-generation pipeline is an unbiased policy-gradient estimator.

For notational simplicity, we suppress the user index when it is clear from context.

\subsection{Proposition 1: Fixed-Candidate Surrogate Gradient}

At turn $t$, let $M_t = \{m_1,\dots,m_K\}$ denote the retrieved memory candidate set, and let $A_t \subseteq M_t$ denote the subset of memories actually injected into the prompt. Let the effective user state be
\begin{equation}
z_t^{\mathrm{eff}} = \beta_L z_t^{(L)} + \beta_S z_t^{(S)}.
\end{equation}
Define the user-aware score
\begin{equation}
s_t(m;z) = \tilde{s}(u_t,m) + \langle z, v_m\rangle,
\end{equation}
where $\tilde{s}(u_t,m)$ collects all score terms that are independent of the user state. The fixed-candidate retrieval policy is
\begin{equation}
\pi_z(m \mid u_t, M_t)
=
\frac{\exp(s_t(m;z)/\tau)}
{\sum_{j:m_j \in M_t}\exp(s_t(m_j;z)/\tau)}.
\end{equation}
For the observed selected set $A_t$, define the surrogate objective
\begin{equation}
\mathcal{J}_t(z)
=
A_t^{\mathrm{retr}}
\cdot
\frac{1}{|A_t|}
\sum_{m \in A_t}
\log \pi_z(m \mid u_t, M_t).
\end{equation}
Also define
\begin{equation}
\begin{aligned}
v_{\mathrm{chosen},t}
&=
\frac{1}{|A_t|}
\sum_{m \in A_t} v_m, \\
\mu_t(z)
&=
\sum_{m \in M_t}
\pi_z(m \mid u_t, M_t)\,v_m.
\end{aligned}
\end{equation}

\textbf{Proposition 1.}
\textit{For any fixed candidate set $M_t$ and observed selected set $A_t$,}
\begin{equation}
\nabla_z \mathcal{J}_t(z)
=
\frac{A_t^{\mathrm{retr}}}{\tau}
\bigl(v_{\mathrm{chosen},t} - \mu_t(z)\bigr).
\end{equation}
\textit{Consequently, evaluated at $z = z_t^{\mathrm{eff}}$, the implemented updates in Section~3.6 satisfy}
\begin{equation}
\begin{aligned}
\Delta z_t^{(L)}
&=
\frac{\eta_L}{\beta_L}\,
\nabla_{z^{(L)}}\mathcal{J}_t\!\left(z_t^{\mathrm{eff}}\right), \\
\Delta z_t^{(S)}
&=
\frac{\eta_S}{\beta_S}\,
\nabla_{z^{(S)}}\mathcal{J}_t\!\left(z_t^{\mathrm{eff}}\right).
\end{aligned}
\end{equation}
\textit{whenever $\beta_L,\beta_S > 0$. Equivalently, after absorbing the fixed factors $\beta_L,\beta_S$ into the learning rates, both implemented updates are exact gradient-ascent steps on the surrogate objective in the effective-state parameterization.}

\textit{Proof.}
For any $m \in M_t$,

\begin{equation}
\begin{aligned}
&\nabla_z \log \pi_z(m \mid u_t, M_t) \\
&= \nabla_z\!\left[
\frac{1}{\tau}\langle z, v_m\rangle
-
\log \sum_{j:m_j \in M_t}
\exp\!\bigl(s_t(m_j;z)/\tau\bigr)
\right] \\
&= \frac{1}{\tau}
\left(
v_m
-
\sum_{j:m_j \in M_t}
\pi_z(m_j \mid u_t, M_t)\,v_{m_j}
\right) \\
&= \frac{1}{\tau}\bigl(v_m - \mu_t(z)\bigr).
\end{aligned}
\end{equation}

Averaging over $m \in A_t$ and multiplying by $A_t^{\mathrm{retr}}$ gives
\begin{align}
\nabla_z \mathcal{J}_t(z)
&=
\frac{A_t^{\mathrm{retr}}}{\tau}
\left(
\frac{1}{|A_t|}
\sum_{m \in A_t} v_m
-
\mu_t(z)
\right) \\
&=
\frac{A_t^{\mathrm{retr}}}{\tau}
\bigl(v_{\mathrm{chosen},t} - \mu_t(z)\bigr).
\end{align}
Finally, since
\begin{equation}
z_t^{\mathrm{eff}} = \beta_L z_t^{(L)} + \beta_S z_t^{(S)},
\end{equation}
the chain rule yields
\begin{equation}
\begin{aligned}
\nabla_{z^{(L)}}\mathcal{J}_t(z_t^{\mathrm{eff}})
= \beta_L \nabla_z \mathcal{J}_t(z_t^{\mathrm{eff}}), \\
\nabla_{z^{(S)}}\mathcal{J}_t(z_t^{\mathrm{eff}})
= \beta_S \nabla_z \mathcal{J}_t(z_t^{\mathrm{eff}}).
\end{aligned}
\end{equation}
which implies the stated identities. \hfill $\square$

\paragraph{Scope.}
Proposition~1 is a statement about the fixed-candidate surrogate induced by the observed candidate set $M_t$, the observed selected set $A_t$, and the scalar weight $A_t^{\mathrm{retr}}$. It does not claim that the full retrieval pipeline with Top-$J$ truncation and heuristic attribution is an unbiased policy-gradient estimator.

\subsection{Proposition 2: Exact Two-Timescale Decomposition}

Recall the update rules from Section~3.6:
\begin{equation}
\begin{aligned}
z_{t+1}^{(L)} &= z_t^{(L)} + \Delta z_t^{(L)}, \\
z_{t+1}^{(S)} &= (1-\lambda)z_t^{(S)} + \Delta z_t^{(S)}, \\
0 &< \lambda \le 1.
\end{aligned}
\end{equation}

For a given session, assume the short-term state is initialized as
\begin{equation}
z_1^{(S)} = 0.
\end{equation}

\textbf{Proposition 2.}
\textit{For all $t \ge 2$, the long-term and short-term states admit the exact unrolled forms}
\begin{equation}
z_t^{(L)}
=
z_1^{(L)} + \sum_{i=1}^{t-1}\Delta z_i^{(L)},
\end{equation}
\begin{equation}
z_t^{(S)}
=
\sum_{i=1}^{t-1}(1-\lambda)^{\,t-1-i}\Delta z_i^{(S)}.
\end{equation}
\textit{Therefore, the effective user state can be written exactly as}
\begin{equation}
\begin{aligned}
z_t^{\mathrm{eff}}
&=
\beta_L z_1^{(L)}
+ \beta_L \sum_{i=1}^{t-1}\Delta z_i^{(L)} \\
&\quad
+ \beta_S \sum_{i=1}^{t-1}(1-\lambda)^{\,t-1-i}\Delta z_i^{(S)} .
\end{aligned}
\end{equation}
\textit{If, in addition, $\|\Delta z_i^{(S)}\| \le G$ for all $i$, then for any $H \ge 0$, the contribution of short-term updates older than $H$ turns is bounded by}
\begin{equation}
\left\|
\sum_{i=1}^{t-1-H}(1-\lambda)^{\,t-1-i}\Delta z_i^{(S)}
\right\|
\le
\frac{G(1-\lambda)^H}{\lambda}.
\end{equation}

\textit{Proof.}
The expression for $z_t^{(L)}$ follows immediately by recursively expanding
\begin{equation}
z_{t+1}^{(L)} = z_t^{(L)} + \Delta z_t^{(L)}.
\end{equation}
Similarly, repeatedly unrolling
\begin{equation}
z_{t+1}^{(S)} = (1-\lambda)z_t^{(S)} + \Delta z_t^{(S)}
\end{equation}
and using $z_1^{(S)} = 0$ gives
\begin{equation}
z_t^{(S)}
=
\sum_{i=1}^{t-1}(1-\lambda)^{\,t-1-i}\Delta z_i^{(S)}.
\end{equation}
Substituting these two identities into
\begin{equation}
z_t^{\mathrm{eff}} = \beta_L z_t^{(L)} + \beta_S z_t^{(S)}
\end{equation}
yields the exact decomposition of the effective state.

For the tail bound, apply the triangle inequality:
\begin{align}
&\left\|
\sum_{i=1}^{t-1-H}(1-\lambda)^{t-1-i}\Delta z_i^{(S)}
\right\| \\
&\le
\sum_{i=1}^{t-1-H}(1-\lambda)^{t-1-i}\|\Delta z_i^{(S)}\| \\
&\le
G \sum_{j=H}^{\infty}(1-\lambda)^j \\
&=
\frac{G(1-\lambda)^H}{\lambda}.
\end{align}

\paragraph{Interpretation.}
Proposition~2 shows that the two state variables operate on different time scales. The long-term state is a persistent accumulator of past update increments, so signal that repeatedly appears across turns can be retained indefinitely. In contrast, the short-term state is an exponentially weighted sum of recent increments, with effective memory horizon on the order of $1/\lambda$. Thus, the dual-vector parameterization separates persistent cross-turn signal from transient local context, without requiring stronger assumptions such as convexity, concavity, or dynamic-regret optimality.

\section{System Mode Descriptions}
\label{app:system-modes}

We compare six system modes under the same frozen-backbone,
no-additional-training setting (rows in
Table~\ref{tab:msc-main}):

\begin{itemize}[nosep,leftmargin=*]
  \item \textbf{Vanilla}: No preference extraction or preference
    memory. The agent conditions only on the current session
    history.

  \item \textbf{Contextual}: The full conversation history is
    appended to the prompt, but no structured preference memory
    or learned user state is maintained.

  \item \textbf{All-memory}: Preference extraction is enabled,
    but all extracted preference notes are appended to the prompt
    at each turn, without retrieval or user-specific selection.

  \item \textbf{Reflection}: After each session, the agent
    generates a reflection summary that is appended to future
    prompts, requiring one additional LLM pass per session.

  \item \textbf{RAG}: Preference cards are constructed, retrieved
    by dense search, reranked, and injected into the prompt. No
    learned user state is used; retrieval depends only on
    query--memory relevance.

  \item \textbf{VARS}: Our full method. This mode uses the
    same preference-memory pipeline as \textbf{RAG}, but augments
    retrieval with the learned user state
    $(z_U^{(L)}, z_{U,t}^{(S)})$ described in
    Section~\ref{sec:method-user-vector}, updated online from
    weak feedback as in Section~\ref{sec:method-rl}.
\end{itemize}

For each system mode, we evaluate the same $60$ user profiles
over the same $60$ sessions and task sequences.

\section{Qualitative Failure Modes}
\label{app:failure-modes}

We observe three recurring failure modes for our RAG-based methods,
particularly as the preference memory grows over sessions:

\begin{enumerate}[nosep]
  \item \textbf{Over-cautious clarification.}
    The agent asks clarification questions instead of
    solving the task when many conflicting preferences are
    retrieved.

  \item \textbf{Preference overload.}
    The agent attempts to satisfy all injected preferences
    simultaneously, producing verbose or unfocused responses.

  \item \textbf{Early user disengagement.}
    The simulator terminates the session when initial responses
    are weak, preventing recovery in subsequent turns.
\end{enumerate}

These failure modes are more pronounced for RAG ($44.3\%$
timeout) than VARS ($26.4\%$), suggesting the user vector
mitigates preference overload by biasing retrieval toward
preferences that have historically led to positive feedback.

\section{Query Transformation for Dense Retrieval}
\label{app:query-transform}

Dense retrieval faces a semantic gap between task-oriented
queries (e.g., ``solve this integral'') and preference
descriptions (e.g., ``when solving math problems, show
step-by-step work''). To reduce this mismatch, we apply a
lightweight keyword-based transformation. Given $u_t$, we detect a
task type (e.g., math, coding, writing, explanation) using
curated keyword lists and, when matched, form a supplementary
retrieval query $u'_t$ by prepending a task-specific prefix such
as \texttt{"user preferences for coding: "} to the original
query. We embed both $u_t$ and $u'_t$, and for each memory card
use the maximum cosine similarity across the two query embeddings
during dense retrieval. The top-$K$ candidates by this score are
passed to the reranker, which uses only the original query
$u_t$. This heuristic is intended to improve recall of
task-conditioned preferences without an additional LLM call.

\section{Reward and Gating Sensitivity Analysis}
\label{app:sensitivity}

Using the interaction logs from the ablation experiments
(Section~\ref{sec:results-ablation}), we re-run user-vector updates
under perturbed reward and gating configurations to assess sensitivity.
For each configuration, we report the final long-term vector
$\ell_2$ norm (averaged over users), the percentage change
relative to the baseline, and the cosine similarity between the
perturbed and baseline long-term vectors.

\paragraph{Reward perturbations.}
Table~\ref{tab:reward-sensitivity} shows that the learned vectors
are robust to moderate reward perturbations: removing half of the
negative keywords or shifting the topic-coherence dampening
threshold has no effect (cosine similarity $= 1.0$). Larger
perturbations---adding noisy negative keywords or tightening the
reward clip range---change both vector magnitude and direction,
indicating that keyword quality matters but the signal is stable
under small noise.

\begin{table}[t]
  \centering
  \small
  \begin{tabularx}{\columnwidth}{Xccc}
    \toprule
    Config & $\|z^{(L)}\|$ & $\Delta\%$ & $\cos$ to base \\
    \midrule
    Baseline             & 0.288 & ---    & 1.000 \\
    Half neg keywords    & 0.288 & 0\%    & 1.000 \\
    Noisy neg keywords   & 0.603 & +109\% & 0.184 \\
    Clip $[-0.5, 0.5]$   & 0.374 & +30\%  & 0.490 \\
    Dampen thresh $0.1$  & 0.288 & 0\%    & 1.000 \\
    Dampen thresh $0.3$  & 0.288 & 0\%    & 1.000 \\
    \bottomrule
  \end{tabularx}
  \caption{Reward sensitivity. Baseline uses the default keyword
    list and clip range $[-1,1]$ with dampening threshold $0.2$.}
  \label{tab:reward-sensitivity}
\end{table}

\paragraph{Gating perturbations.}
Table~\ref{tab:gating-sensitivity} shows that the retrieval-attribution
gate is critical for stable vector learning. Removing gating entirely
($g_t = 1.0$) causes $454\%$ vector norm inflation and severe
directional drift ($\cos = 0.41$), confirming that ungated reward
noise propagates into the user state. A uniform gate ($g_t = 0.5$)
partially mitigates this but still inflates vectors by $168\%$.
In contrast, shifting the similarity thresholds used in the gating
logic has no effect, indicating that the gating mechanism is robust
to threshold choice within a reasonable range.

\begin{table}[t]
  \centering
  \small
  \caption{Gating sensitivity. Baseline uses the heuristic gating
    with similarity thresholds $(0.2, 0.5)$.}
  \label{tab:gating-sensitivity}
  \begin{tabularx}{\columnwidth}{Xccc}
    \toprule
    Config & $\|z^{(L)}\|$ & $\Delta\%$ & $\cos$ to base \\
    \midrule
    Fixed $g = 0.5$       & 0.771 & +168\% & 0.444 \\
    Fixed $g = 1.0$       & 1.593 & +454\% & 0.410 \\
    Thresh $(0.1, 0.4)$   & 0.288 & 0\%    & 1.000 \\
    Thresh $(0.3, 0.6)$   & 0.288 & 0\%    & 1.000 \\
    \bottomrule
  \end{tabularx}
\end{table}

\section{Memory Card Schema and Example}
\label{app:card-example}

Each memory card stored in the preference memory has the following fields:

\begin{lstlisting}[style=jsonprofile]
{
  "id": "8475ca85-...",
  "note": "When presenting calculations, omit extra
           phrases and show only the essential steps.",
  "condition": "presenting calculations",
  "action": "omit extra phrases, show essential steps",
  "is_global": false,
  "embedding": "[4096-dim dense vector from f_emb]",
  "item_vec": "[256-dim PCA projection v_m]"
}
\end{lstlisting}

The \texttt{condition} and \texttt{action} fields are produced by the preference extractor $M_{\text{ext}}$ (Section~\ref{sec:method-extractor}). The \texttt{note} field is a natural-language summary used as input to the reranker $M_{\text{rerank}}$. The \texttt{is\_global} flag determines whether the preference bypasses retrieval and is injected directly into the prompt (Section~\ref{sec:method-memory}). The \texttt{embedding} and \texttt{item\_vec} fields are computed from the source user query as described in Section~\ref{sec:method-memory}.

\end{document}